\documentclass[letterpaper, 10 pt, conference]{ieeeconf}  

\IEEEoverridecommandlockouts                              
\usepackage{times}
\usepackage{multirow}
\usepackage{epsfig}
\usepackage{graphicx}
\usepackage{color}
\usepackage{amsmath}
\usepackage{amssymb} 
\usepackage{amsfonts}
\usepackage{rotating}
\usepackage{makecell}
\usepackage{tabularx}
\usepackage{subfig}
\usepackage{mathtools}
\usepackage{graphics} 
\usepackage{stackrel}
\usepackage{siunitx}
\usepackage[table,xcdraw]{xcolor}
\usepackage[noline,noend,ruled]{algorithm2e}
\usepackage[normalem]{ulem}
\usepackage{booktabs}

\newcommand{\etal}{\textit{et al}. }

\newcolumntype{C}[1]{>{\centering\arraybackslash}p{#1}}

\graphicspath{{figs/}{../figs/}}

\DeclareGraphicsExtensions{.eps,.jpg,.png}

\definecolor{airforceblue}{rgb}{0.36, 0.54, 0.66}
\definecolor{atomictangerine}{rgb}{1.0, 0.6, 0.4}
\definecolor{critical}{rgb}{1.0, 0.0, 0.0}
\definecolor{comment}{rgb}{0.0, 0.0, 1.0}
\definecolor{comment_chr}{rgb}{0.0, 0.4, 0.0}

\newcommand{\Xk}{\mathbf{X}_k}
\newcommand{\Fk}{\mathbf{F}_k}
\newcommand{\Ek}{\mathbf{E}_k}
\newcommand{\Ik}{\mathbf{I}_k}
\newcommand{\Pc}{\mathbf{P}}
\newcommand{\Ep}{\mathbf{E}_P}
\newcommand{\Pk}{\mathbf{P}_k}
\newcommand{\Dk}{\mathbf{D}_k}

\title{\LARGE \bf
LAPTNet: LiDAR-Aided Perspective Transform Network}

\author{ Manuel Diaz-Zapata$^{1,2}$, \"{O}zg\"{u}r Erkent$^{1,3}$, Christian Laugier$^{1}$, Jilles Dibangoye$^{1,2}$, David Sierra-Gonzalez$^{1}$   
\thanks{This work was partially supported by EU project CPS4EU and Toyota Motor Europe.}
\thanks{ 
$^{1}$ Authors are with the Chroma team, INRIA Grenoble Rhone-Alpes, France. $^{2}$ Authors are with CITI-Lab, INSA Lyon, France. $^{3}$ Author is with Hacettepe University, Ankara, Turkey.
Correspondence: {\tt\small manuel.diaz-zapata@inria.fr}}%
}

\begin{document}

\maketitle
\thispagestyle{empty}
\pagestyle{empty}

\begin{abstract}

\noindent Semantic grids are a useful representation of the environment around a robot. They can be used in autonomous vehicles to concisely represent the scene around the car, capturing vital information for downstream tasks like navigation or collision assessment. Information from different sensors can be used to generate these grids. Some methods rely only on RGB images, whereas others choose to incorporate information from other sensors, such as radar or LiDAR. In this paper, we present an architecture that fuses LiDAR and camera information to generate semantic grids. By using the 3D information from a LiDAR point cloud, the LiDAR-Aided Perspective Transform Network (LAPTNet) is able to associate features in the camera plane to the bird's eye view without having to predict any depth information about the scene. Compared to state-of-the-art camera-only methods, LAPTNet achieves an improvement of up to 8.8 points (or 38.13\%) over state-of-art competing approaches for the classes proposed in the NuScenes dataset validation split.

\end{abstract}

\section{Introduction}
 \label{sec:intro}





For an autonomous vehicle, sensing its surroundings is a crucial task. To do this, the vehicle can make use of an array of different sensors, such as cameras or LiDARs, to gather information about the environment that it is in. Cameras have been one of the most widely used sensors for tasks such as image segmentation, 2D and 3D object detection \cite{Zhao-2017-CVPR,redmon2018yolov3,caddn}. LiDARs have also been used to a lower extent for 3D point cloud segmentation \cite{pointnet} and 3D object detection \cite{pointpillars}. Other works have also explored the fusion of both types of sensors for tasks like detection with great success \cite{fpointnet}. 

\begin{figure}
\centering
\includegraphics[width=0.9\columnwidth]{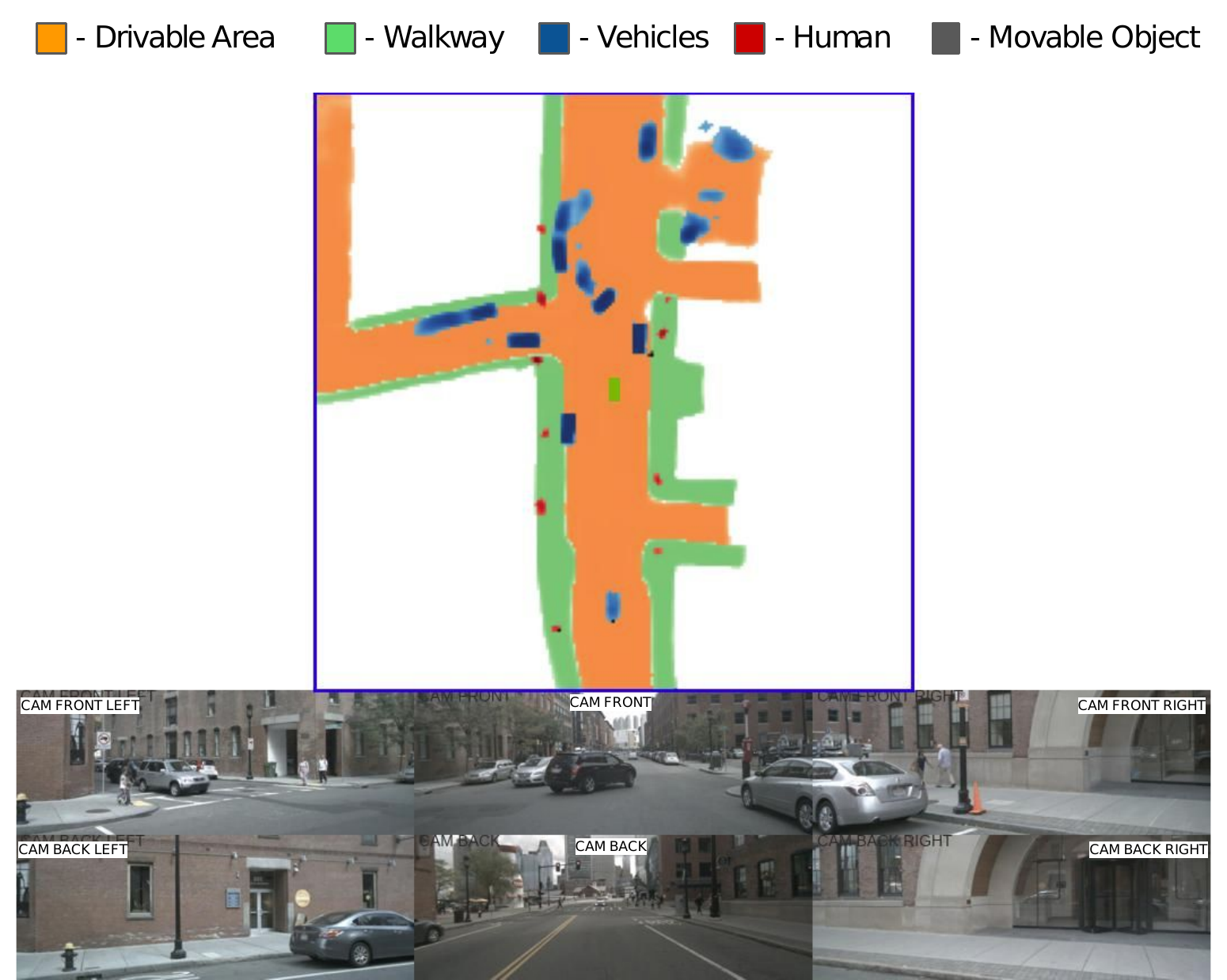}
\caption{\label{fig:overview} Semantic grid predictions by LAPTNet for classes in the NuScenes dataset \cite{nuscenes}. By using LIDAR information to aid the projection of camera features, our method allows the creation of more precise semantic grids compared to camera-only methods. Best viewed with digital zoom.}
\vspace{-6mm}
\end{figure}


Many methods have used a camera as the only sensor. They usually address tasks that are performed in the camera plane, like semantic segmentation \cite{Zhao-2017-CVPR} or object detection \cite{fcos}. These works, although important to the field of scene understanding, rely on a representation space that suffers from perspective distortions \cite{cv-book}. This can result in differences of the final output for objects with similar sizes in the real world, e.g. bigger bounding boxes or segmentation masks for cars closer to the autonomous vehicle than for those farther away.

With multi-layer LiDAR sensors, perspective distortion is not an issue since the points are already in 3D. Here, the problem is related to the density of the information and the structure in which they will be processed. Compared to a camera, the amount of data points given by LiDAR is very sparse, which can result in the loss of important details present in the scene. 


In robotics, occupancy grids are a representation space that can be quickly generated from 3D information \cite{cmcdot}. They usually represent a discretized version of the top-down view, sometimes called bird's-eye view (BEV), which is usually the plane in which a robot can move. For this, a 3D point cloud can be used to generate a 2D array that indicates which cells are occupied or not. In comparison, using a monocular image to build this representation space using only geometrical methods, like inverse perspective mapping \cite{cv-book}, is very challenging. This is due to the uncertainty of the pinhole model for determining the depth correspondence for each pixel in the image \cite{cv-book}. This problem can be alleviated by using more than one camera at the same time, as in stereo depth estimation, or by using feature matching methods for a moving camera, as in structure from motion \cite{cv-book}.   

Recently, the problem of how to generate semantic grids has started to receive lots of attention from the community working on perception for autonomous vehicles \cite{Erkent-2020-RAL,roddick-pon, pillarsegnet}. Here, the sensor that has received most of the attention has been the camera. Semantic occupancy grids have been estimated by leveraging stereo depth \cite{hoyer-semgrids} or by using neural networks to learn implicit depth distributions \cite{philion-lss}, to go from the camera plane to the BEV plane. The fusion of camera and LiDAR information to generate semantic grids has also been explored by other works \cite{Erkent-2020-RAL, deepfusion, transfuser} using different approaches.

In this paper, we propose a fusion architecture that leverages the 3D information from LiDAR together with the features extracted from camera images to generate semantic grids. The proposed approach uses the information from the point cloud to guide the projection of features extracted from camera images, performing the association between the image plane and the bird's-eye view representation in a fast and efficient manner without having to rely on estimation methods to project image information to the BEV. We address the sparse association problem between camera pixels and LiDAR points by leveraging the downsampling feature from convolutional neural networks (CNNs) to do the point-pixel correlation. In fig. \ref{fig:overview}, the resulting semantic grids can be seen with the corresponding set of surround images used to generate them. 

In Section ~\ref{sec:rellit}, we present the related literature; in Section ~\ref{sec:method}, we explain our network architecture, and how LiDAR information helps to project the image features onto the BEV to generate semantic grids; in Section ~\ref{sec:exp_setup}, we describe our experimental setup on the NuScenes Dataset and in Section ~\ref{sec:results} we present our results on the NuScenes dataset with 5 different classes. Finally, we summarize the findings of our work.\vspace{-3.2mm}

\section{Related Work}
\label{sec:rellit}

In this section we will present some of the current works in the literature which use the bird's eye view space for different perception tasks and how it is generated from camera or LiDAR data. We will begin by presenting some works on 3D bounding box detection in subsection \ref{subsec:bev3dbbox}, followed by methods that estimate semantic grids through camera-based methods, LiDAR-based methods and fusion-based methods in subsection \ref{subsec:semgrids}.

\vspace{-1mm}

\subsection{Using the Bird's Eye View for bounding box detection}
\label{subsec:bev3dbbox}
Recent works have been interested in using the BEV space as an intermediary representation space for the prediction of 3D and 2D bounding boxes from either camera or LiDAR information. Camera-only methods perform the projection from the camera plane to the BEV using different approaches. OFT uses an intermediary voxel space to associate image features to the BEV \cite{OFT}. A Generative Adversarial Network \cite{goodfellow2014generative} is used to do the projection in \cite{2d-3d-lifting}. In CaDDN, the camera features are projected to 3D via categorical depth distributions that are predicted with a Frustum Feature Network, they are later collapsed to the BEV via convolutions \cite{caddn}. 

Aside from camera-based models, other methods like Voxel R-CNN use LiDAR, where a BEV representation is generated by stacking in the Z axis the voxels generated from the point cloud that are processed by a 3D network \cite{voxel-rcnn}. Deep Continous Fusion \cite{deep-cont-fusion} proposes to use the BEV as the space to fuse information from cameras and LiDAR through a 'Continous Fusion Layer' based on \cite{dpcc}. \vspace{-1.5mm}

\subsection{Semantic Grid estimation}
\label{subsec:semgrids}
The bird's eye view representation space can be better related to the task of estimating semantic grids than the estimation of 3D bounding boxes. To create them, cameras and LiDARs are two of the most used sensors. In this subsection, will present how semantic grids have been created using only camera or LiDAR data, as well as some sensor fusion approaches. 

\subsubsection{Using only cameras}
In the Pyramid Occupancy Network the proposed dense transformers project each of the output scales from an FPN \cite{fpn}-inspired backbone to the feature map in the BEV \cite{roddick-pon}. Lift-splat-shoot tackles the projection problem by jointly predicting a set of categorical depths together with the image features from a 2D backbone, then performing voxel pooling to generate the BEV used for predicting the semantic grid \cite{philion-lss} . In the VPN, a set of camera viewpoints are projected to the BEV using a View Transformer Module, which is a multilayer perceptron that finds the relationship between each pixel in each view to each cell in the BEV \cite{vpn}. Yang \etal use attention mechanisms to project extracted features from monocular images to the BEV, similar to the VPN \cite{project-vpn}. Hoyer \etal project semantic labels from the image plane to the BEV using stereo depth information \cite{hoyer-semgrids}. 

For the camera-base approaches, the pinhole model does not allow to project from the 2D image to a certain 3D point without extra depth information. These methods address this shortcoming in different ways by using stereo depth estimation, learning depth distributions or by using an intermediary bottleneck dimension to go from the camera plane to BEV. These estimation approaches present inherent uncertainties that affect the precision of the models for the creation of semantic grids.

\begin{figure*}[t]
   \centering
   \includegraphics[width=0.8\linewidth]{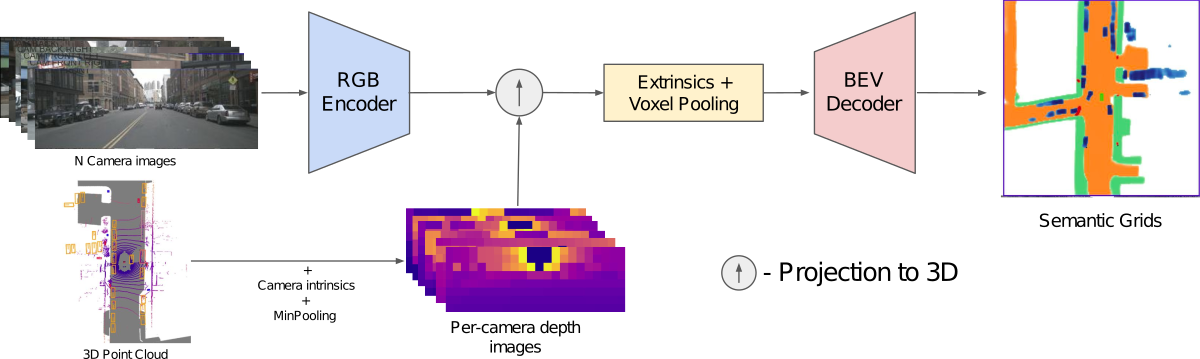}
   \caption{Proposed architecture for the LAPTNet. We encode per-camera features using a CNN on the images. These features are then projected to 3D using their corresponding depth images. These depth images are generated by projecting the LiDAR point cloud to each of the cameras' field of view using the intrinsics of the cameras together with a minpooling operation to match the downsampling performed by the CNN. The points in 3D are aligned to the vehicle's reference frame and projected to the BEV through a voxel sum-pooling operation \cite{philion-lss}. Finally, a BEV decoder network generates the semantic grids from this BEV feature map.}
   \label{fig:arch}
   \vspace{-5mm}
\end{figure*}

\subsubsection{Using only LiDAR}

PillarSegNet \cite{pillarsegnet} extends the work of \cite{pointpillars} to generate semantic grids by using a combination of Pillar features with an occupancy feature map (both in the BEV). CMCDOT \cite{cmcdot} estimates static and dynamic occupancy states, as well as empty and unknown states for cells in a grid using a Bayesian filtering approach.

Although LiDAR approaches have the advantage of working with 3D information from point clouds, they do not offer the density of information, nor the texture information available in RGB images that can be beneficial to understand how the scene is composed. 

\subsubsection{Fusing different sensors}
Erkent \etal propose the late fusion of camera and LiDAR data on the BEV through an encoder-decoder network to predict the semantic grid. Semantic segmentation images are projected via inverse perspective mapping to a set of intermediary planes which are then concatenated to CMCDOT grids to be fused together \cite{erkent2019end}. 
FISHINGNet does late fusion for information from cameras, LiDAR and radar to predict semantic grids. Camera features are projected here using the VPN \cite{vpn}. Each modality is processed separately to predict semantic grids, they are then fused via pooling to generate the final output \cite{hendy2020fishing}. 

These methods rely on camera-based approaches for the projection of image features to the BEV, as well as separate pipelines for LiDAR and camera data. An early fusion scheme could allow sensors to cover each others shortcomings earlier, and their information to be jointly processed for faster inference.   


\section{LAPTNet: LIDAR-Aided Perspective Transform Network}
\label{sec:method}
In this section we describe how our approach, the LIDAR-Aided Perspective Transform Network (LAPTNet) generates semantic grids by using 3D point clouds to guide the projection of camera features to the BEV. A general overview of the method can be seen in Fig. \ref{fig:arch}. \vspace{-1mm}


\subsection{Problem formulation}
\label{sec:method-subsec:prob_form}

We are given $n$ images $\{\Xk \in \mathbb{R}^{3 \times H \times W}\}_{n}$ taken from the cameras located around a vehicle, each with a corresponding intrinsic ($\Ik \in \mathbb{R}^{3 \times 3}$) and extrinsic ($\mathbf{E_{k}} \in \mathbb{R}^{4 \times 4}$) camera matrix. We are also given a point cloud ($\Pc \in \mathbb{R}^{3 \times D}$) taken by the LiDAR at the same time, with its corresponding transformation matrix ($\Ep \in \mathbb{R}^{4 \times 4}$) from the vehicle's reference frame. 

Using this information, we want to estimate an occupancy grid ($\mathbf{y} \in \mathbb{R}^{C \times X \times Y}$) in the BEV centered on the coordinate frame of the vehicle. Here, by leveraging the 3D geometric information available in $\Pc$, we can project features from $\Xk$ to the BEV without having to deal with the depth estimation task for each pixel.

\begin{table*}[t]
 \label{table:general_results}
   \centering
\begin{tabular}{@{}lccccc@{}}
\toprule
                                             & Human           & Vehicle         & Movable Object   & Drivable Area    & Walkway          \\ \midrule
\multicolumn{1}{l|}{VPN \cite{vpn}}                     & 7.1\%           & 13.47\%         & 7.7\%            & 58.0\%           & 29.4\%           \\
\multicolumn{1}{l|}{PON \cite{roddick-pon}}                     & 8.2\%           & 15.37\%         & 6.9\%            & 60.4\%           & 31.0\%           \\ 
\multicolumn{1}{l|}{Lift-Splat-Shoot \cite{philion-lss}}        & 9.99\%          & 32.02\%         & 21.6\%           & 77.6\%           & 51.03\%          \\\midrule
\multicolumn{1}{l|}{FISHINGNet (LiDAR and Camera) \cite{hendy2020fishing}}              & \textbf{20.4\%} & \textbf{40.9\%} & -                & -                & -                \\
\multicolumn{1}{l|}{\textbf{LAPTNet (Ours)}} & 13.8\%          & 40.13\%         & \textbf{27.45\%} & \textbf{79.43\%} & \textbf{57.25\%} \\ \bottomrule
\end{tabular}
\caption{Results on the NuScenes validation split. We perform the comparison of the Intersection over Union of the generated semantic grids. Best results are presented in bold font.}
\label{tab:results_general}
\end{table*}

\begin{table*}[t]
   \label{table:rain}
   \centering
\begin{tabular}{@{}lccccc@{}}
\toprule
                                             & Human            & Vehicle          & Movable Object   & Drivable Area    & Walkway          \\ \midrule
\multicolumn{1}{l|}{Lift-Splat-Shoot (Rain)}        & 5.16\%           & 33.2\%           & 27.15\%          & 71.95\%          & 47.05\%          \\ 
\multicolumn{1}{l|}{\textbf{LAPTNet} (Rain)} & \textbf{10.05\%} & \textbf{44.76\%} & \textbf{32.37\%} & \textbf{73.07\%} & \textbf{49.96\%} \\ \midrule
\multicolumn{1}{l|}{Lift-Splat-Shoot (Night)}        & 4.99\%          & 31.44\%         & 4.85\%          & 64.47\%          & 22.93\%         \\ 
\multicolumn{1}{l|}{\textbf{LAPTNet} (Night)}  & \textbf{6.29\%} & \textbf{36.8\%} & \textbf{12.3\%} & \textbf{67.88\%} & \textbf{25.9\%} \\ \bottomrule
\end{tabular}
\caption{Results on the NuScenes validation scenes under rain (top) and night (bottom) conditions.}
\label{tab:results_rain}
\vspace{-5mm}
\end{table*}

\subsection{Processing of camera images}
\label{sec:method-subsec:cnn_prepro}

We want to have as many correspondences between points in $P$ and pixels in $\Xk$ to generate a representation in the BEV. But given the sparsity of data in point clouds compared to the pixel density in images, if we directly project the point cloud to the image, a big amount of information will be lost since not all pixels in the original image will get a point correspondence to perform the projection to the BEV.

Instead of projecting the RGB values from the original image, we choose to preprocess $\Xk$ with CNNs. Following the standard CNN encoding, we are able to downsample each $\Xk$ by a factor of $d_f$ to a feature map $\{\Fk  \in \mathbb{R}^{N_{f} \times H/d_f \times W/d_f}\}_n$ containing $N_f$ channel-wise features. This allows us to perform the association in a smaller space with the possibility of finding a higher ratio of point-pixel correspondences. 

\subsection{Projection of camera features onto the BEV plane}
\label{sec:method-subsec:lapt}

The main idea behind LAPTNet is to leverage the geometrical information about the scene, encoded in $\Pc$, to associate the pixels in $\Xk$ from the camera plane to their corresponding cells in the intermediary BEV representation $(\mathbf{B} \in \mathbb{R}^{N_f \times X \times Y})$. By projecting $\Pc$ to the field of view of $\Xk$ we can associate image features with their corresponding point in 3D where each pixel is looking to. 

For this, we transform $\Pc$ from the LiDAR reference frame to the camera reference frame using the camera's extrinsics $(\Ek)$ and the LiDAR transformation matrix ($\Ep$). The transformation is done using homogeneous coordinates as indicated in equation \ref{eq:transformLidartoCam}.\vspace{-1mm}

\begin{equation}
\label{eq:transformLidartoCam}
    \begin{pmatrix} \Pk \\ 1 \end{pmatrix} = \Ek \times \Ep^{-1} \times \begin{pmatrix} \Pc \\ 1  \end{pmatrix} 
\end{equation}

With the points now in the camera reference frame $(\Pk = (x_{k}, y_{k}, z_{k})^T)$, a perspective transformation from 3D space to the 2D image coordinates $(u_k,v_k)$ is performed using the camera's intrinsic matrix $(\Ik)$. As can be seen in equation \ref{eq:transform3Dto2D}, we normalize by the value in the depth dimension $(z_{k})$ for each point in $\Pk$.

\begin{equation}
\label{eq:transform3Dto2D}
    \begin{pmatrix}
        u_k \\
        v_k \\
        1
    \end{pmatrix} = \Ik \times \begin{pmatrix}
        x_k/z_{k} \\
        y_k/z_{k} \\
        z_{k}/z_{k}
    \end{pmatrix}
\end{equation}

Knowing where the points from $\Pc$ are projected to $\Xk$, we only keep the point closest to the camera coordinate frame in the $z_k$ axis. We choose the closest point since we are using the pinhole camera projection model \cite{cv-book}. With this information, we create a sparse depth image $\{\Dk \in \mathbb{R}^{1 \times H \times W}\}_n$ for each camera, saving the depth as the pixel value where the points are projected to.

Finally, in order to use the information in $\Dk$ to project $\Fk$ to the BEV, we need to reduce the depth image's dimensions to match the feature map. By performing a minpooling operation with a kernel of size $d_f$ we find the closest distance value in the receptive field for each pixel of $\Fk$ in $\Dk$. An example of this low-resolution depth map can be seen in Fig. \ref{fig:arch}.  These distance values $\{\delta_k \in \mathbb{R}^{1\times H/d_f \times W/d_f}\}_n$ are used to do the projection to 3D in the camera's reference frame as shown in equation \ref{eq:transform2Dto3D}. A feature located in coordinates $(u_f, v_f)^T$ in $\Fk$, will be projected to the point $(x_f, y_f, z_f)^T$ in 3D space.
\vspace{-3.2mm}

\begin{equation}
\label{eq:transform2Dto3D}
    \begin{pmatrix}
        x_f \\
        y_f \\
        z_f
    \end{pmatrix}_k = \Ik^{-1} \times \delta_k \times \begin{pmatrix}
        u_f \\
        v_f \\
        1
    \end{pmatrix}_k
\end{equation}\vspace{-3.2mm}

\begin{equation}
\label{eq:transform_TFs}
    \begin{pmatrix}
        x_f \\
        y_f \\
        z_f \\
        1
    \end{pmatrix}_{car} = \Ek^{-1} \times \begin{pmatrix}
        x_f \\
        y_f \\
        z_f \\
        1
    \end{pmatrix}_k
\end{equation}

 With the point cloud containing the features' position in 3D, we transform them from the camera's reference frame to the vehicle's reference frame (equation \ref{eq:transform_TFs}). Then, we perform the projection to the BEV by following the voxel pooling method described in \cite{philion-lss}. We create the intermediary representation $\mathbf{B}$ by assigning every point $(x_f,y_f,z_f)_{car}^T$ to its nearest pillar and performing sum pooling in each pillar. Here, "pillars" refer to voxels with infinite height as described in \cite{pointpillars}. \vspace{-3mm}

\subsection{Semantic grid generation using the projected features}

Since the representation space $\mathbf{B}$ follows the same structure as an image, we use a lightweight CNN as the BEV decoder that outputs the semantic grid. This top-down network is a ResNet-18 network that ends with 2 upsampling blocks (bilinear upsampling operation, (3x3) convolution, batch normalization and ReLU) to recover the original grid spatial size $(X,Y)$ after passing through the decoder. A final (1x1) convolution predicts the wanted output  ($\mathbf{y} \in \mathbb{R}^{C \times X \times Y}$) with $C$ channels for each cell in the grid.\vspace{-2mm}

\begin{figure*}[t]
   \centering
   \includegraphics[width=0.7\linewidth]{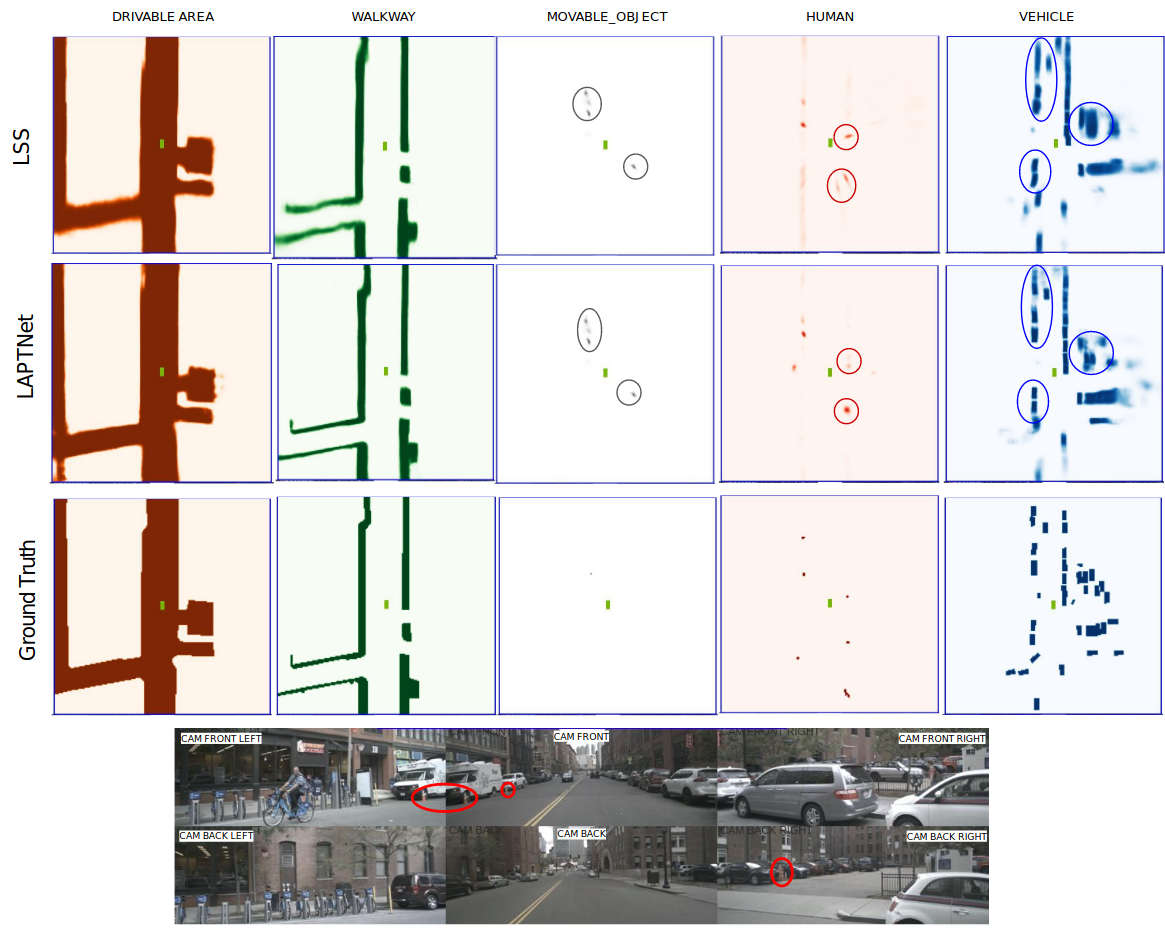}
   \caption{Comparison of the predicted semantic grids by Lift-Splat-Shoot (LSS) \cite{philion-lss} and LAPTNet. For some of the classes whose objects occupy a few cells in the grid, like human or movable object, LAPTNet seems to predict a more precise location than LSS. The red circles in the images highlight some movable objects that are not present on the ground truth annotations and are detected by both methods (third column). For classes such as vehicles, movable objects and humans the predictions from LSS are stretched along the projection ray, this does not happen in the LAPTNet predictions. Best viewed with digital zoom.}
   \label{fig:comparison}
   \vspace{-3mm}
\end{figure*}

\section{Experimental Setup}
\label{sec:exp_setup}
In this section we will discuss the experimental setup for the training of LAPTNet. We will describe the dataset, loss function and metric used for the evaluation of model performance under general and difficult conditions. \vspace{-2mm}

\subsection{Dataset}
We base all of our experiments on the NuScenes dataset \cite{nuscenes}. NuScenes is a large dataset with 1000 driving scenes from different locations around the world. Each scene of the dataset has a duration of 20 seconds, recording information from a variety of sensors such as LiDAR, cameras and radars. Since the semantic grid ground truth is not directly available from the dataset, we extend the method of \cite{philion-lss} to generate the ground truth for all of the chosen classes. Using the 3D bounding box annotations and the high-definition maps available in the dataset, we generate our ground truth semantic grids for the different classes. Given the annotation scheme followed by NuScenes, we decide to group the 3D bounding box annotations into the classes 'human', 'movable object' and 'vehicle' as well as taking the 'drivable area' and 'walkway' classes from the HD map.


\subsection{Loss function and Evaluation metrics}
\vspace{-1mm}We train our network separately for each class $(C=1)$. Knowing this, we employ the binary cross entropy loss as our loss function with a weight for positive samples equal to 2.13. We chose this value given the implementation of \cite{philion-lss}. We use the Intersection over Union (IoU) metric to evaluate how similar the predicted segmentation masks are to the ground truth. 
\vspace{-2mm}

\subsection{Competing approaches}
To the extent of our knowledge, no other method apart from \cite{hendy2020fishing} generates semantic grids using a sensor fusion approach in the NuScenes dataset. Knowing this, we compare our method to current state of the art baselines that use only camera-based approaches such as \cite{philion-lss,roddick-pon,vpn}.

We also report the performance of our model under specific conditions such as those of rain and night. These two conditions are of interest to us given the effect that rain can have in the accuracy of LiDAR as well as low-light for cameras. \vspace{-3mm}

\section{Results}
\label{sec:results}

In this section we discuss the results of our approach, how it compares to some of the current state of the art methods and how our method compares under conditions such as night and rain.

\subsection{Comparison against state of the art}
\vspace{-0.8mm}
We report our quantitative results on Table \ref{tab:results_general}. In this table, we can observe that by adding the LiDAR information we are able to outperform all of the camera-only baselines. We hypothesize that this improvement is due to the addition of the LiDAR data, since the real depth information available in it informs where objects are located in space better than prediction-based baselines. 

We report improvements of \textbf{3.8} points (38.13\%) in the human metaclass, of \textbf{8.8} points (25.33\%) in the vehicle metaclass, of \textbf{5.85} points (27.08\%) in the movable object metaclass, of \textbf{1.83} points (2.35\%) in the drivable area class and of \textbf{6.22} points (12.18\%) in the walkway class over the highest performing camera-only method (LSS \cite{philion-lss}). 

We also observe that even without adding a LiDAR-specific encoder, we get close results to FISHINGNet \cite{hendy2020fishing} for the vehicle class. A noticeable difference can be seen in the performance for the human class, showing potential for improving the performance of LAPTNet. By adding a feature extractor specific for LiDAR to be fused with the projected camera features, we expect to improve the performance of the network. We also interpret these results as another indicator of the overall value generated from adding the real 3D information from the LiDAR sensor. 

A sample of the qualitative results compared against the best competing method can be seen in Fig. \ref{fig:comparison}. Here it can be seen that LAPTNet manages a more concise assignment of cells that belong to classes with a smaller footprint such as human and movable object. In contrast for LSS, the prediction masks are streched along the projection ray used to reach those cells. For vehicles, LAPTNet is able to more precisely assign the cells to which the vehicles belong to. We also note the interesting case for the two traffic cones highlighted by a red circle in the camera images of Fig. \ref{fig:comparison}, where even if they are not annotated in the ground truth, both LAPTNet and LSS seem to have detected them. 

\subsection{Performance under difficult conditions}

Since we do not rely on only one type of sensor for our prediction, we expect our method to be robust under adverse conditions such as rain or night. The results for the scenes under rain condition can be seen in the upper part of table \ref{tab:results_rain} and the results for night condition can be seen in the lower part. 

Comparing against the highest-performing camera-based model, we observe that our method outperforms the baseline in both types of adverse conditions. In this study we find further support to the idea that not only the color or texture features taken from the camera images are important to generate the semantic grids, but the geometric distribution of them within the grid space (given by the LiDAR) is what enables our method to outperform the baselines. 

Looking at the results for both rain and night, we think that our method still seems to rely more on the camera features than the geometric distribution of them along the grid. This idea stems from the fact that cameras perform worse under low-light situations, and since the projected features in the BEV come from a camera-only encoder, even adding the depth information from the LiDAR is not enough to fully overcome this challenge. On the other hand, when comparing the performance under rainy conditions, a bigger challenge for LiDARs than for cameras, we see that the drop in segmentation accuracy is not as big as it is for night conditions. We have reason to think that, as previously stated, adding a parallel branch in the style of \cite{pointpillars} can help us improve the results in the night condition since LiDAR is not affected by this type of adversity.


\section{Conclusion and Future Work}
\label{sec:conclusion}
\noindent
We have presented here the novel approach of the LIDAR-Aided Projective Transform Network (LAPTNet). This network uses real geometric information from a LiDAR sensor to guide the projection of camera features onto a bird's eye view perspective for semantic occupancy grid generation. Our method consistently improves the grid segmentation performance in the classes defined for the nuScenes dataset.

As future work, we plan to evaluate how the performance of the model changes when using multiple feature scales. We also plan to evaluate if the performance changes when the LiDAR information is processed with a separate encoder.



\section*{Acknowledgement}
\noindent
\small
The experiments presented in this paper were carried out using the Grid'5000 testbed, supported by a scientific interest group hosted by Inria and including CNRS, RENATER and several Universities as well as other organizations.

\bibliographystyle{IEEEtran}
\bibliography{egbib}   

\end{document}